\documentclass[11pt,a4paper]{article}
\usepackage[hyperref]{acl2020}
\usepackage{times}
\usepackage{latexsym}

\usepackage{verbatim}
\usepackage{pifont}
\usepackage{multirow}
\usepackage{amsmath}
\usepackage{capt-of}
\usepackage{tabularx}
\usepackage[caption=false]{subfig}
\usepackage{epsfig}
\usepackage{amssymb}
\usepackage{amsfonts}
\usepackage{booktabs}
\usepackage{scalerel}
\usepackage[inline]{enumitem}
\usepackage{listings}
\usepackage{varwidth}
\usepackage[export]{adjustbox}
\usepackage{tikz}
\usetikzlibrary{tikzmark}
\usepackage{todonotes}
\usepackage{cleveref}

\usepackage{stmaryrd}
\usepackage{bbm}

\usepackage{amsmath,amsfonts,bm}

\def\eqref#1{equation~\ref{#1}}
\def\1{\bm{1}}

\DeclareMathAlphabet{\mathsfit}{\encodingdefault}{\sfdefault}{m}{sl}
\SetMathAlphabet{\mathsfit}{bold}{\encodingdefault}{\sfdefault}{bx}{n}

\usepackage[ruled,noend]{algorithm2e}
\usepackage{graphicx}
\usepackage{url}

\usepackage{footnote}
\makesavenoteenv{table}

\aclfinalcopy % Uncomment this line for the final submission
\def\aclpaperid{812} %  Enter the acl Paper ID here

\newcommand\refqa{\textsc{RefQA}}

\newcommand{\cmark}{\ding{51}}%
\newcommand{\xmark}{\ding{55}}%

\newcommand{\linote}[1]{{%
  \let\thempfn\relax% Remove footnote number printing mechanism
  \footnotetext[0]{\emph{#1}}% Print footnote text
}}

\title{Harvesting and Refining Question-Answer Pairs \\ for Unsupervised QA}
\author{Zhongli Li$^\dagger$\thanks{~~Contribution during internship at Microsoft Research.},~~Wenhui Wang$^\ddagger$,~~Li Dong$^\ddagger$,~~Furu Wei$^\ddagger$,~~Ke Xu$^\dagger$ \\
  $^\dagger$Beihang University \\
  $^\ddagger$Microsoft Research \\
  \texttt{\{lizhongli@,kexu@nlsde.\}buaa.edu.cn} \\
  \texttt{\{wenwan,lidong1,fuwei\}@microsoft.com}}

\date{}

\begin{document}
\maketitle
\begin{abstract}
Question Answering (QA) has shown great success thanks to the availability of large-scale datasets and the effectiveness of neural models. Recent research works have attempted to extend these successes to the settings with few or no labeled data available.
In this work, we introduce two approaches to improve unsupervised QA. 
First, we harvest lexically and syntactically divergent questions from Wikipedia to automatically construct a corpus of question-answer pairs (named as \refqa{}).
Second, we take advantage of the QA model to extract more appropriate answers, which iteratively refines data over \refqa{}.
We conduct experiments\footnote{The code and data are available at \url{https://github.com/Neutralzz/RefQA}.} on SQuAD 1.1, and NewsQA by fine-tuning BERT without access to manually annotated data.
Our approach outperforms previous unsupervised approaches by a large margin and is competitive with early supervised models.
We also show the effectiveness of our approach in the few-shot learning setting.
\end{abstract}

\section{Introduction}

% \begin{figure}[t]
%     \centering
%     \includegraphics[width=7.5cm]{images/intro-example2.pdf}
%     \caption{An overview of our approach. \refqa{} Construction (top) is constructing a dataset using the statements with references in Wikipedia. Iterative Data Refinement (bottom) is using QA model to help extract answers and generate new data.}
%     \label{fig:ex-intro}
% \end{figure}

Extractive question answering aims to extract a span from the given document to answer the question. 
Rapid progress has been made because of the release of large-scale annotated datasets~\citep{squad1, squad2, triviaqa}, and well-designed neural models~\citep{wang2016matchlstm, bidaf, qanet}. %(such as Match-LSTM~\citep{wang2016matchlstm}, BiDAF~\citep{bidaf}, DCN~\citep{dcn}, R-Net~\citep{wang-rnet}, DrQA~\citep{drqa}, QANet~\citep{qanet}). 
Recently, unsupervised pre-training of language models on large corpora, such as BERT~\citep{BERT}, has brought further performance gains.

% Supervised approaches have achieved great progress for this task in the past years. In particular, pretrained language model, e.g. BERT~\citep{bert}, XLNet~\citep{xlnet} and Roberta~\citep{roberta}, have significantly matched human performance on SQuAD benchmark~\citep{squad1,squad2}.

However, the above approaches heavily rely on the availability of large-scale datasets. The collection of high-quality training data is time-consuming and requires significant resources, especially for new domains or languages.
In order to tackle the setting in which no training data available,~\citet{lewis2019unsupervisedqa} 
%propose to generate natural question-answer pairs in an unsupervised manner.
leverage unsupervised machine translation to generate synthetic context-question-answer triples.
%\citet{yang-semiqa} and~\citet{dhingra2019semi} make good progress requiring only a small set of labeled examples in a semi-supervised setting.
% Recently,~\citet{lewis2019unsupervisedqa} have explored the unsupervised method, which creates synthetic question-answer-context triples in a unsupervised manner.
The paragraphs are sampled from Wikipedia. NER and noun chunkers are employed to identify answer candidates.
Cloze questions are first extracted from the sentences of the paragraph, and then translated into natural questions. 
However, there are a lot of lexical overlaps between the generated questions and the paragraph. 
Similar lexical and syntactic structures render the QA model tend to predict the answer just by word matching.
Moreover, the answer category is limited to the named entity or noun phrase, which restricts the coverage of the learnt model.
% extracting answer candidates by NER or noun chunkers limits the categories of answers.

% the generated questions have a lot of overlaps with the paragraph, given the cloze questions are extracted from 
% generating questions from the sentence or clause of the paragraph
% generate cloze questions and generate questions from the sentence or clause of paragraph using unsupervised machine translation. 
%But the diversity of questions is restricted by their approach because: 

\begin{figure*}[t]
    \centering
    \includegraphics[width=16cm]{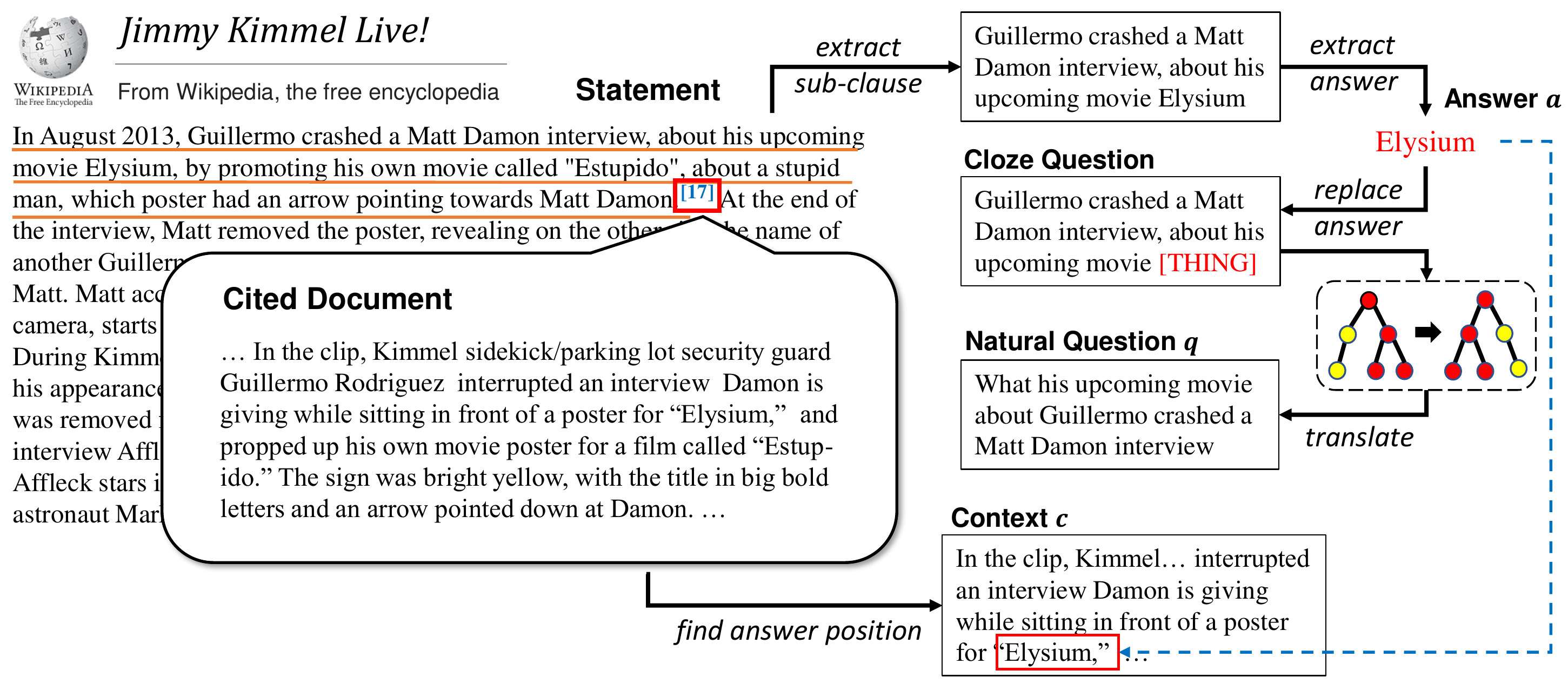}
    \caption{Overview of \refqa{} construction.}
    \label{fig:da}
\end{figure*}

In this work, we present two approaches to improve the quality of synthetic context-question-answer triples. 
First, we introduce the \refqa{} dataset, 
%a question answering dataset constructed in an unsupervised manner, 
which harvests lexically and syntactically divergent questions from Wikipedia by using the cited documents. 
As shown in Figure~\ref{fig:da}, the sentence (statement) in Wikipedia and its cited documents are semantically consistent, but written with different expressions. 
More informative context-question-answer triples can be created by using the cited document as the context paragraph and extracting questions from the statement in Wikipedia.
Second, we propose to iteratively refine data over~\refqa{}.
% , which uses a BERT-based QA model to extract appropriate and diverse answer candidates, and generates questions for these answers to construct the refined data.
%As shown in Figure~\ref{fig:data_refinement}, 
Given a QA model and some~\refqa{} examples, we first filter its predicted answers with a probability threshold. Then we refine questions based on the predicted answers, and obtain the refined question-answer pairs to continue the model training.
Thanks to the pretrained linguistic knowledge in the BERT-based QA model, there are more appropriate and diverse answer candidates in the filtered predictions, some of which do not appear in the candidates extracted by NER tools.
%We generate questions for new answer candidates, obtain the refined question-answer pairs and combine them with filtered data to continue the model training. 
We also show that iteratively refining the data further improves model performance.

We conduct experiments on SQuAD 1.1~\citep{squad1}, and NewsQA~\citep{newsqa2017}. Our method yields state-of-the-art results against strong baselines in the unsupervised setting. 
Specifically, the proposed model achieves 71.4 F1 on the SQuAD 1.1 test set and 45.1 F1 on the NewsQA test set without using annotated data. We also evaluate our method in a few-shot learning setting. Our approach achieves 79.4 F1 on the SQuAD 1.1 dev set with only 100 labeled examples, compared to 63.0 F1 using the method of~\citet{lewis2019unsupervisedqa}.

% Our approach significantly outperforms previous state-of-the-art methods with a considerable margin.
% Meanwhile, experimental results show that the two proposed methods improve performance consistently.
% We also conduct a few-shot learning task with limited labelled training examples, and our approach with only 1000 labeled examples achieve 85.2 F1 score on SQuAD 1.1 dev set.
%Our QA model with data augmentation (single model) achieves a 66.8 F1 score, including absolute gains of 12.1 than the previous state-of-the-art result~\citep{lewis2019unsupervisedqa}. Based on the performance, our question-regeneration method finally improves the F1 score to 70.7, amounting to 3.9 absolute improvement. 
%We also demonstrate that our syntax-guided translation method shows competitive results on \refqa, comparing to other unsupervised question generation methods~\citep{lewis2019unsupervisedqa}.

To summarize, the contributions of this paper include: 
i) \refqa{}~constructing in an unsupervised manner, which contains more informative context-question-answer triples.
% that harvested from Wikipedia and constructed by an unsupervised manner; 
ii) Using the QA model to iteratively refine and augment the question-answer pairs in \refqa{}.
% a novel data refinement to further encourage QA model's own prediction, which help us extract more diverse and appropriate answers. 
%and (iii) a syntax-guided translation that puts the answer-related words in front, improving the performance with a considerable margin.

\section{Related Work}

\paragraph{Extractive Question Answering}
Given a document and question, the task is to predict a continuous sub-span of the document to answer the question. 
% The answers are constrained to be a continuous sub-span of the document in EQA. 
Extractive question answering has garnered a lot of attention over the past few years. 
Benchmark datasets, such as SQuAD~\citep{squad1,squad2}, NewsQA~\citep{newsqa2017} and TriviaQA~\citep{triviaqa}, play an important role in the progress.
In order to improve the performance on these benchmarks, several models have been proposed, including BiDAF~\citep{bidaf}, R-NET~\citep{wang-rnet}, and QANet~\citep{qanet}. 
Recently, unsupervised pre-training of language models such as BERT~\citep{BERT}, 
% which learns general language/world knowledge from a large text corpus by simply training to predict word tokens based on its surrounding context, 
achieves significant improvement. 
However, these powerful models rely on the availability of human-labeled data. Large annotated corpora for a specific domain or language are limited and expensive to construct.

\paragraph{Semi-Supervised QA} 
Several semi-supervised approaches have been proposed to utilize unlabeled data.
Neural question generation (QG) models are used to generate questions from unlabeled passages for training QA models~\cite{yang-semiqa,unans-qg,syntheticqa,unilm}. However, the methods require labeled data to train the sequence-to-sequence QG model. 
\citet{dhingra2019semi} propose to collect synthetic context-question-answer triples by generating cloze-style questions from the Wikipedia summary paragraphs in an unsupervised manner. 
%~\citet{yang-semiqa} follow the GAN~\citep{goodfellow2014generative} paradigm, using QG as a generator and QA as a discriminator, to augment the labeled data with generated questions to train the QA model.
%~\citet{dhingra2019semi} collect large-scaled cloze-style QA pairs to pretrain a model, then improve the performance by fine-tuning on labeled data. 
%~\citet{wang-etal-2018-multi} improve semi-supervised cloze-style QA by multi-perspective aggregation. 

\paragraph{Unsupervised QA} \citet{lewis2019unsupervisedqa} have explored the unsupervised method for QA. They create synthetic QA data in four steps. i) Sample paragraphs from the English Wikipedia. ii) Use NER or noun chunkers to extract answer candidates from the context. iii) Extract ``fill-in-the-blank'' cloze-style questions given the candidate answer and context. iv) Translate cloze-style questions into natural questions by an unsupervised translator. Compared with~\citet{dhingra2019semi},~\citet{lewis2019unsupervisedqa} attempt to generate natural questions by training an unsupervised neural machine translation (NMT) model. They train the NMT model on non-aligned corpora of natural questions and cloze questions. The unsupervised QA model of~\citet{lewis2019unsupervisedqa} achieves promising results, even outperforms early supervised models.
% They create synthetic context-question-answer triples to train QA models, which are generated automatically from unlabeled data. They sample paragraphs from English Wikipedia, and use NER or noun chunkers to extract answer candidates from context. Next they extract the sentences containing answers from context, replace answers with blanks to get the ``fill-in-the-blank" cloze questions and borrow the idea of unsupervised machine translation~\citep{lample2017unsupervised, lample2018phrase, xlm} to perform cloze-to-natural-question translation.
However, their questions are generated from the sentences or sub-clauses of the same paragraphs, which may lead to a biased learning of word matching since its similar lexicons and syntactic structures. 
Besides, the category of answer candidates is limited to named entity or noun phrase, which restricts the coverage of the learnt QA model. 
% Besides, they only focus the answer categories on named entity or noun phrase, and rely on pretrained components to extract answers entirely, that causes the extracted span in context maybe not suitable as a answer.

\section{Harvesting \refqa{} from Wikipedia}
\label{sec:da}

%Given a question $q$ and a paragraph $c$, Extractive QA aims to extract an answer $a$ with starting position $b$ and end position $e$ in the paragraph. For unsupervised QA, we need to construct the triple $(c, q, a)$ data to train QA systems.

In this section, we introduce \refqa{}, a question answering dataset constructed in an unsupervised manner.
One drawback of~\citet{lewis2019unsupervisedqa} is that questions are produced from the paragraph sentence that contains the answer candidate. So there are considerable expression overlaps between generated questions and context paragraphs.
In contrast, we harvest informative questions by taking advantage of Wikipedia's reference links, where lexical and syntactic differences exist between the article and its cited documents.

% \refqa{} is our augmented QA dataset that constructed in an unsupervised manner.
% Instead of generating questions from context in~\citet{lewis2019unsupervisedqa}, we harvest divergent questions by taking advantage of the lexical difference between statements in Wikipedia and corresponding references.
%to prevent a lot of overlaps between questions and context. 
% We elaborate the constructing process in the following section.
%In this section, to prevent a lot of overlaps between questions and context, we propose an approach of constructing synthetic QA data by taking advantage of the difference in expression between statements in Wikipedia and corresponding references, instead of generating question from context.

As shown in Figure~\ref{fig:da}, given statements in Wikipedia paragraphs and its cited documents, we use the cited documents as the context paragraphs and generate questions from the sub-clauses of statements.
In order to generate question-answer pairs, we first find answer candidates that appear in both sub-clauses and context paragraphs. 
Next, we convert sub-clauses into the cloze questions based on the candidate answers.
We then conduct cloze-to-natural-question translation by a dependency tree reconstruction algorithm.
We describe the details as follows.

% Figure~\ref{fig:da}  shows an example of our constructing process.
% We first obtain the statement in Wikipedia and the cited document. Then we treat the cited document as the context paragraph, and use the Wikipedia statement to extract answer and generate question, which is more divergent from the context.

\subsection{Context and Answer Generation}
Statements in Wikipedia and its cited documents often have similar content, but are written with different expressions.
Informative questions can be obtained by taking the cited document as the context paragraph, and generate questions from the statement.
We crawl statements with reference links from the English Wikipedia.
The cited documents are obtained by parsing the contents of reference webpages.

% We only consider the sources of references from namely web pages, newspaper articles, and press.

% Our context are taken from the references of the Wikipedia articles.
%The statements with references are crawled from the English Wikipedia. 
%Through hyperlinks of references, we obtain the context paragraphs by parsing the HTML contents.
% We only consider the sources of references from namely web pages, newspaper articles, and press. Then we obtain the cited documents by parsing the HTML contents into plain text through hyperlinks of references. 

%We treat the cited documents as our context paragraphs and use the Wikipedia statements for our answer span extraction and question generation.

% \subsection{Answer Extraction}
%Given the context $c$ and the statements $s$, we need to create answer spans $a$ which appear both in $c$ and $s$.  
% Answers are extracted from the Wikipedia statements. Initially, we focus answer spans on named entities using a NER tool, here requires supervision, but no aligned QA data. Then we use the QA model to extract more appropriate and diverse answers in Section~\ref{sec:data_refinement}.
Given a statement and its cited document, we restrict the statement to its sub-clauses, and extract answer candidates (i.e., named entities) that appear in both of them by using a NER toolkit.
We then find the answer span positions in the context paragraph. If the candidate answer appears multiple times in the context, we select the position whose surrounding context has the most overlap with the statement.
% we divide the context into sentences and 
% Given context paragraph and a answer candidate, we 
% To find the appropriate answer position in the context, we divide the context into sentences and get the answer position
% sentence containing the answer. 
%If the answer is not in context, we discard it. 
% If there are multiple sentences in context containing the answer, we select the sentence which overlaps most with the  statement. 
% Then we consider the position where the answer first appears in the sentence as our answer position.

\subsection{Question Generation}
\label{ssec:qg_section}

We first generate cloze questions~\cite{lewis2019unsupervisedqa} from the sub-clauses of Wikipedia statements.
Then we introduce a rule-based method to rewrite them to more natural questions, which utilizes the dependency structures.
% We translate cloze questions to natural questions by a simple heuristic method, which moves answer-related words in the dependency parsing tree to the front of the question.
% Moreover, questions asked by human start with answer-related words in most cases, so we translate clozes by a dependency tree reconstruction method that move answer-related words to front.
%, instead of using unsupervised machine translation.

\subsubsection{Cloze Generation}
Cloze questions are the statements with the answer replaced to a mask token.
Following~\citet{lewis2019unsupervisedqa}, we replace answers in statements with a special mask token, which depends on its answer category\footnote{We obtain the answer type labels by a NER toolkit, and group these labels to high-level answer categories, which are used as our mask tokens, e.g., \texttt{PRODUCT} corresponding to \texttt{THING}, \texttt{LOC} corresponding to \texttt{PLACE}.}.
Using the statement and the answer (with a type label \texttt{PRODUCT}) from Figure~\ref{fig:da}, this leaves us with the cloze question ``\textit{Guillermo crashed a Matt Damon interview, about his upcoming movie }\texttt{[THING]}".

% \begin{table}[t]
%     \centering
%     \begin{tabular}{ll}
%         \toprule
%         Answer Type  & Question Word\\ \midrule
%         PERSON/NORP/ORG\footnote{PERSON: people, including fictional; NORP: nationalities or religious or political groups; ORG: companies, agencies, institutions, etc.} & Who \\
%         PLACE & Where \\
%         THING & What  \\
%         TEMPORAL & When \\
%         NUMERIC & How many/much \\
%         \bottomrule
%     \end{tabular}
%     \caption{The grouped answer types and corresponding question words.}
%     \label{tab:qmap}
% \end{table}

% We replace answers in statements with our grouped answer types which shown in Table~\ref{tab:qmap}. 
%The extracted named entity has an type label and we group this label into five categories as shown in Table~\ref{tab:qmap}. 
% Assume the statement ``\textit{Guillermo crashed a Matt Damon interview , about his upcoming movie Elysium}" and the answer ``\textit{Elysium}", we replace the answer ``\textit{Elysium}" with the mask token ``\textit{THING}" to get the cloze question.

\subsubsection{Translate Clozes to Natural Questions}
\label{sec:dep:rec}
%To make questions more relevant to answers, 

%To generate natural questions, we first apply identity mapping to the cloze questions. Following~\citet{lewis2019unsupervisedqa}, we replace the mask token with a wh* word using a simple heuristic rule.

We perform a dependency reconstruction to generate natural questions. We move answer-related words in the dependency tree to the front of the question, since answer-related words are important.
The intuition is that natural questions usually start with question words and question focus~\cite{qa:as:ie}.

\begin{figure}[t]
    \centering
    \includegraphics[width=\linewidth]{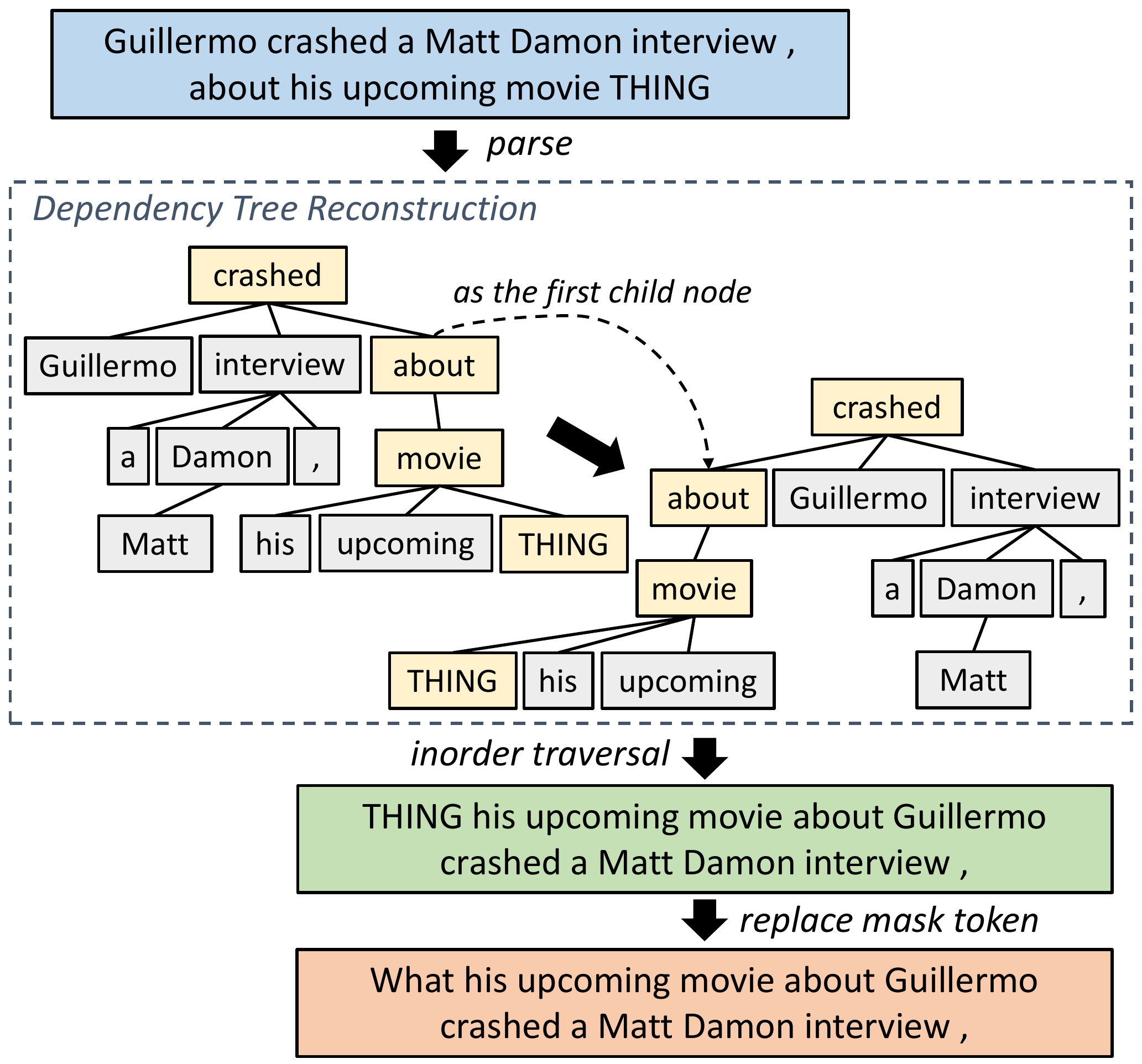}
    \caption{Example of translating cloze questions to natural questions. The node with light yellow color indicates that its subtree contains the answer node.}
    \label{fig:translate_cloze}
\end{figure}

As shown in Figure~\ref{fig:translate_cloze}, we apply the dependency parsing to the cloze questions, and translate them to natural questions by three steps: i) We keep the right child nodes of the answer and prune its lefts. ii) For each node in the parsing tree, if the subtree of its child node contains the answer node, we move the child node to the first child node.
iii) Finally, we obtain the natural question by inorder traversal on the reconstructed tree. We apply the same rule-based mapping as~\citet{lewis2019unsupervisedqa}, which replaces each answer category with the most appropriate wh* word. For example, the \texttt{THING} category is mapped to ``\textit{What}".

% as shown in Figure~\ref{fig:translate_cloze}.We first apply the dependency parsing to clozes, and then change the structure of the dependency tree by the following scheme:
%finally replace the mask token to question word through the map shown in Table \ref{tab:qmap}. 
% (i) if the tree node is the answer node (node text is the answer), we keep its right child nodes and discard its lefts; (ii) if the subtree of the node contains answer node, we set the child node that contains answer node as the first child node. Finally, we replace the mask token to question word through the map shown in Table \ref{tab:qmap}, and the question can be obtained by inorder traversal on the reconstructed tree. 
%We recursively implement this scheme and the pseudo code is presented in Algorithm \ref{alg:tree}.

\begin{comment}
\begin{algorithm}
\caption{Dependency Reconstruction}\label{alg:tree}
\begin{algorithmic}[1]
    \Function{Reconstruction}{$node$}
    \If {$node$ is $answer~node$}
        \State $\text{the left child nodes of } node \gets \phi$
        \State \Return true
    \EndIf
    \State $L\text{ is the list of children in }node$
    \For {each $child$ in $L$}
            \If {\Call{Reconstruction}{$child$}}
                \State $\text{move }child\text{ to the front of }L$
                \State \Return true
            \EndIf
    \EndFor
    \State \Return false
    \EndFunction
\end{algorithmic}
\end{algorithm}
\end{comment}

\section{Iterative Data Refinement}
\label{sec:data_refinement}

\begin{figure}[t]
\centering
\includegraphics[width=\linewidth]{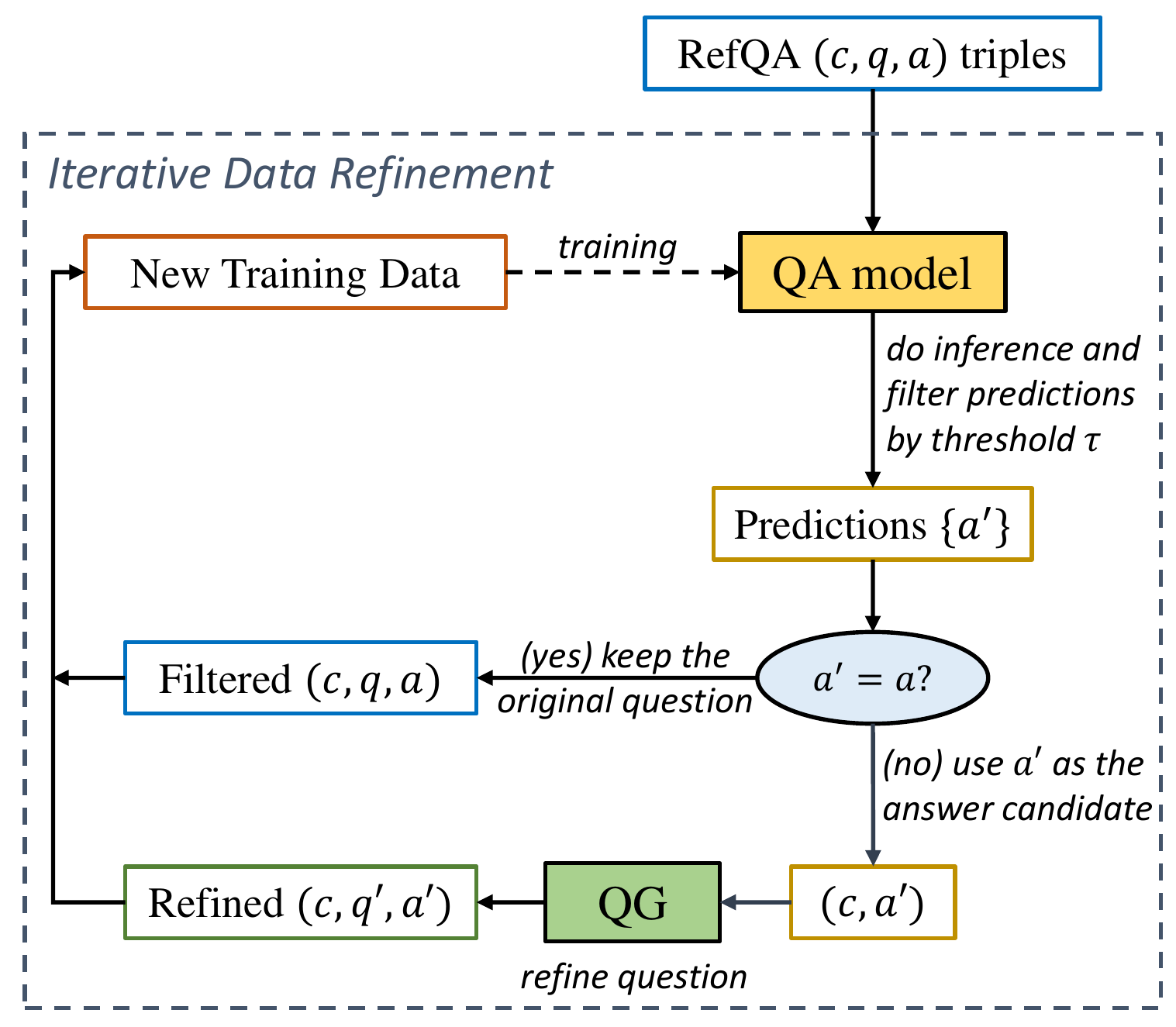}
\caption{Overview of our iterative data refinement process. ``QG" is the process of question generation as described in Section~\ref{ssec:qg_section}. We produce new training data and iteratively train the QA model.}
\label{fig:data_refinement}
\end{figure}

In this section, we propose to iteratively refine data over \refqa{} based on the QA model. 
As shown in Figure~\ref{fig:data_refinement}, we use the QA model to filter \refqa{} data, find appropriate and diverse answer candidates, and use these answers to refine and augment \refqa{} examples.
Filtering data can get rid of some noisy examples in \refqa{}, and pretrained linguistic knowledge in the BERT-based QA model finds more appropriate and diverse answers. We produce questions for the refined answers, then continue to train the QA model on the refined and filtered triples.

\subsection{Initial QA Model Training}
The first step of iterative data refinement is to train an initial QA model.
We use the \refqa{} examples $S_I=\{(c_i,q_i,a_i)\}_{i=1}^{N}$ to train a BERT-based QA model $P(a|c,q)$ by maximizing:
\begin{equation}
\sum_{S_I}{\log~P(a_i|c_i,q_i)}
\end{equation}
where the triple consists of context $c_i$, question $q_i$, and answer $a_i$.

\subsection{Refine Question-Answer Pairs}
As shown in Figure~\ref{fig:data_refinement}, the QA model $P(a|c,q)$ is used to refine the \refqa{} examples. 
We first conduct inference on the unseen data (denoted as $S_U$), and obtain the predicted answers and their probabilities.  
Then we filter the predicted answers with a confidence threshold $\tau$:
\begin{equation}
\label{eq:answer_set}
\begin{split}
Z_A = \{a^{\prime}_i |  P(a^{\prime}_i|c_i,q_i) \ge \tau \}_{(c_i,q_i,a_i) \in S_U} \nonumber
\end{split}
\end{equation}
where $a^{\prime}_i$ represents the predicted answer.

For each predicted answer $a^{\prime}_i$, if it agrees with the gold answer $a_i$, we keep the original question.
For the case that $a^{\prime}_i \ne a_i$, we treat $a^{\prime}_i$ as our new answer candidate. Besides, we use the question generator (Section~\ref{ssec:qg_section}) to refine the original question $q_i$ to $q^{\prime}_i$.
% \begin{equation}
%     q^{\prime}_i = \mathop{\arg\max}_{q} P(q|c_i,a^{\prime}_i).
% \end{equation}

In this step, using the QA model for filtering helps us get rid of some noisy examples. The refined question-answer pairs $(q^{\prime}_i,a^{\prime}_i)$ can also augment the \refqa{} examples.
The pretrained linguistic knowledge in the BERT-based QA model is supposed to find more novel answers, i.e., some candidate answers are not extracted by the NER toolkit.
With the refined answer spans, we then use the question generator to produce their corresponding questions.

\subsection{Iterative QA Model Training}

After refining the dataset, we concatenate them with the filtered examples whose candidate answers agree with the predictions.
The new training set is then used to continue to train the QA model.
The training objective is defined as:
\begin{equation}
\begin{split}
    \max & \sum_{a^{\prime}_i \in Z_A} [\mathbb{I}(a^{\prime}_i=a_i)\text{log}~P(a_i|c_i,q_i)\\
        & +\mathbb{I}(a^{\prime}_i \ne a_i)\text{log}~P(a^{\prime}_i|c_i,q^{\prime}_i)],
    %L(\theta_A) =  - & \sum_{x_i \in S_U}  [\text{log}~P(a_i|c_i,q_i;\theta_{A})\\
    %& +\text{log}~P(a^{\prime}_i|c_i,q^{\prime}_i;\theta_{A})]
\end{split}
\end{equation}
where $\mathbb{I}(\cdot)$ is an indicator function (i.e., $1$ if the condition is true).
%This step can be seen that we encourage our model to believe its own prediction is also an appropriate answer.

Using the resulting QA model, we further refine question-answer pairs and repeat the training procedure.
The process is repeated until the performance plateaus, or no new data available.
%Then the optimal parameters can be an alternative ${\theta}_A$ in Equation~\ref{eq:answer_set} to help us refine new data and train the QA model repeatedly until the performance has not changed or no new data available. 
%\wang{Then we use the QA model to refine question-answer pairs again, and update $\theta_A$ iteratively until the performance has not changed or no new data available.}
Besides, in order to obtain more diverse answers during iterative training, we apply a decay factor $\gamma$ for the threshold $\tau$.
The pseudo code of iterative data refinement is presented in Algorithm~\ref{alg:refine}.

%\subsection{Learning Method}
% Answers need respect to questions, but~\citet{lewis2019unsupervisedqa} extract answers from context directly and rely on pretrained components entirely.

\begin{algorithm}[t]
%\small
%\SetAlCapNameFnt{\small}
%\SetAlCapFnt{\small}
\SetAlgoNoLine
\DontPrintSemicolon
\KwIn{synthetic context-question-answer triples $\mathcal{S} = \{(c_i,q_i,a_i)\}_{i=1}^{N}$, a threshold $\tau$ and a decay factor $\gamma$. }
%\KwOut{the question answering model parameters $\theta_A$}
Sample a part of triples $\mathcal{S}_I$ from $\mathcal{S}$\;
Update the model parameters by \\ ~~~~maximizing $\sum_{\mathcal{S}_I} \log P(a|c,q)$\;
Split unseen triples into $\{\mathcal{S}_{U_1},\mathcal{S}_{U_2},...,\mathcal{S}_{U_M}\}$\;
\For{$k \gets 1$ \textbf{to} $M$} {
    $\mathcal{D} \gets \phi$\;
    \For {$(c_i,q_i,a_i)$ \textbf{in} $\mathcal{S}_{U_k}$}{
        $Z_A \gets \{a^{\prime}_i ~~\text{s.t.}~~ P(a^{\prime}_i|c_i,q_i) \ge \tau\}$\;
        \For {$a^{\prime}_i$ \textbf{in} $Z_A$}{
            \uIf{$a^{\prime}_i = a_i$}{
                $\mathcal{D} \gets \mathcal{D} \cup (c_i,q_i,a_i)$\;
            }
            \Else{
                Refine question $q_i$ to $q^{\prime}_i$\;
                %$q^{\prime}_i = \mathop{\arg\max}_{q} P(q|c_i,a^{\prime}_i;\theta_G)$\;
                $\mathcal{D} \gets \mathcal{D} \cup (c_i,q^{\prime}_i,a^{\prime}_i)$\;
            }
        }
    }
    $\tau \gets \tau \times \gamma$\;
    Update the model parameters by \\ ~~~~maximizing $\sum_{\mathcal{D}} \log P(a|c,q)$\;
}
\KwOut{the updated QA model $P(a|c,q)$}
%\Return{$\theta_A$}
\caption{Iterative Data Refinement}
\label{alg:refine}
\end{algorithm}

\section{Experiments}

We evaluate our proposed method on two widely used extractive QA datasets~\citep{squad1,newsqa2017}.
% We conduct ablation studies to understand the different contributions.
We also demonstrate the effectiveness of our approach in the few-shot learning setting.
%Implementation details and experimental results are provided in the following sub-sections.
%In this section, \wang{we present our implementation details for unsupervised QA}, then we report experimental results of the proposed methods on SQuAD 1.1 and NewsQA, finally we conduct an ablation study to understand the different contributions. We also demonstrate the effectiveness of our approach in a few-shot learning setting. 

\begin{table*}[t]
    \centering
    \begin{tabular}{lcccc}
        \toprule
        & \multicolumn{2}{c}{SQuAD 1.1} & \multicolumn{2}{c}{NewsQA} \\
        Models & Dev Set & Test Set & Dev Set & Test Set \\ \midrule
        \multicolumn{5}{l}{\emph{Supervised Methods}} \\
        % \quad LR~\citep{squad1}  & 40.0 / 51.0 & 40.4 / 51.0  & - / - & - / - \\
        \quad DCR~\citep{yu2016dcr} & 62.5 / 71.2 & 62.5 / 71.0  & - / - & - / - \\
        \quad mLSTM~\citep{wang2016matchlstm} & 64.1 / 73.9 & 64.7 / 73.7 & ~~34.4 / 49.6$^*$ & ~~34.9 / 50.0$^*$ \\
        % FastQA~\citep{fastqa2017} & 67.8 / 76.3 & 68.4 / 77.1 & 43.7 / 56.4 & 41.9 / 55.7 \\
        \quad FastQAExt~\citep{fastqa2017} & 70.3 / 78.5 & 70.8 / 78.9 & 43.7 / 56.1 & 42.8 / 56.1 \\
        \quad R-NET~\citep{wang-rnet} & 71.1 / 79.5 &  71.3 / 79.7  & - / - & - / - \\
        \quad BERT-Large~\citep{BERT} & 84.2 / 91.1 &  85.1 / 91.8  & - / - & - / - \\ 
        \quad SpanBERT~\citep{spanbert} & - / - & 88.8 / 94.6 & - / - & ~~~~~~- / 73.6 \\
        \midrule
        \multicolumn{5}{l}{\emph{Unsupervised Methods}} \\
        \quad~\citet{dhingra2019semi}$^{\dagger}$ & 28.4 / 35.8 & - / - & - / - & - / -\\
        \quad~\citet{lewis2019unsupervisedqa} & - / - & 44.2 / 54.7 & - / - & - / - \\
        \quad~\citet{lewis2019unsupervisedqa}$^{\ddagger}$ & 45.4 / 55.6 & - / - & 19.6 / 28.5 & 17.9 / 27.0 \\
        \quad Our \refqa{}  & 57.1 / 66.8 & 55.8 / 65.5 & 29.0 / 42.2 & 27.6 / 41.0 \\
        \quad ~~~~+ Iterative Data Refinement & \textbf{62.5} / \textbf{72.6} & \textbf{61.1} / \textbf{71.4} & \textbf{33.6} / \textbf{46.3} & \textbf{32.1} / \textbf{45.1}\\
        \bottomrule
    \end{tabular}
    \caption{Results (EM / F1) of our method, various baselines and supervised models on SQuAD 1.1, and NewsQA. ``$*$" means results taken from~\citet{newsqa2017}, ``$\dagger$" means results taken from~\citet{lewis2019unsupervisedqa}, and ``$\ddagger$" means our reimplementation on BERT-Large (Whole Word Masking).}
    \label{tab:mainresult}
\end{table*}

\subsection{Configuration}

\paragraph{\refqa{} Construction} 
We collect the statements with references from English Wikipedia following the procedure in~\cite{wikiref}.
We only consider the references that are HTML pages, which results in 1.4M statement-document pairs.
% For each statement, only the first reference is used in the case of multiple references.
%The cited documents are collected from namely web pages, newspaper articles and press.

In order to make sure the statement is relevant to the cited document, we tokenize the text, remove stop words and discard the examples if more than half of the statement tokens are not in the cited document. 
The article length is limited to 1,000 words for cited documents.
Besides, we compute ROUGE-2~\citep{lin-2004-rouge} as correlation scores between statements and context.
We use the score's median ($0.2013$) as a threshold, i.e., half of the data with lower scores are discarded.
We obtain 303K remaining data to construct our \refqa{}.

We extract named entities as our answer candidates, using the NER toolkit of Spacy. We split the statements into sub-clauses with Berkeley Neural Parser~\citep{Kitaev-2018-SelfAttentive}.
The questions are generated as in Section~\ref{ssec:qg_section}.
% , since there is no specific answer entity typing for noun phrases.
% because of the missing type tag for the extracted noun phrases.  
%\wang{Moreover, the question length can affect the performance of QA system. There may be multiple sentences in a statement, it is too long as a question. Thus, before question generation, we split the text to sub-clauses with Berkeley Neural Parser~\citep{Kitaev-2018-SelfAttentive}.}
%, and save the clause that contains the special mask token. 
We also discard sub-clauses that are less than $6$ tokens, to prevent losing too much information of original sentences.
Finally, we obtain 0.9M \refqa{} examples.
% we get 900k examples as our training data \refqa.

\paragraph{Question Answering Model}
% We adopt the uncased version of BERT\footnote{\url{https://github.com/huggingface/pytorch-transformers}} to initialize our QA model and fine-tune on our \refqa{}.
We adopt BERT as the backbone of our QA model.
Following~\cite{BERT}, we represent the question and passage as a single packed sequence.
We apply a linear layer to compute the probability of each token being the start or end of an answer span.
We use Adam~\citep{adam} as our optimizer with a learning rate of 3e-5 and a batch size of 24.
The max sequence length is set to 384.
We split the long document into multiple windows with a stride of 128. 
% Our best performing setup is used to fine-tune a BERT-Large (Whole Word Masking) model.
We use the uncased version of BERT-Large (Whole Word Masking). 
We evaluate on the dev set every 1000 training steps, and conduct early stopping when the performance plateaus. 
%For BERT-base, we use a single P100 GPU to fine-tune \zhongli{and do evaluation after 2 training epochs}.% For a fair comparison, \wang{we train 2 epochs and then do evaluation for different settings on our ablation study.}

\paragraph{Iterative Data Refinement}
We uniformly sample 300k data from \refqa{} to train the initial QA model.
We split the remaining 600k data into 6 parts for iterative data refinement.
For each part, we use the current QA model to refine question-answer pairs. We combine the refined data with filtered data in a 1:1 ratio to continue training the QA model.
Specially, we keep the original answer if its prediction is a part of the original answer during inference.
%\wang{refine answers with questions} and combine the refined data with original data in a 1:1 ratio as the \wang{next training data}. 
The threshold $\tau$ is set to 0.15 for filtering the model predictions. The decay factor $\gamma$ is set to 0.9. 
% Specially, if answer prediction is a part of original answer during inference, we keep the original answer, because we \zhongli{consider} that \zhongli{the named entities are indivisible}.

\subsection{Results}
We conduct evaluation on the SQuAD 1.1~\citep{squad1}, and the NewsQA~\citep{newsqa2017} datasets.
We compare our proposed approach with previous unsupervised approaches and several supervised models. 
Performance is measured via the standard Exact Match (EM) and F1 metrics.
%\wang{The Exact Match (EM) and F1 score are two evaluation metrics for the QA performance}.

%~\citet{squad1} is a simple logistic regression model to find answers by word overlap.

%~\citet{qanet} is an architecture combining convolution with self-attention that achieves better performance than recurrent neural networks.

\citet{dhingra2019semi} propose to train the QA model on the cloze-style questions. Here we take the unsupervised results that re-implemented by~\citet{lewis2019unsupervisedqa} with BERT-Large.
The other unsupervised QA system~\cite{lewis2019unsupervisedqa} borrows the idea of unsupervised machine translation~\citep{lample2017unsupervised} to convert cloze questions into natural questions.
%fine-tune BERT-Large using only Wikipedia data and borrow the idea of unsupervised machine translation~\citep{lample2017unsupervised}  to . 
For a fair comparison, we use their published data\footnote{\url{https://github.com/facebookresearch/UnsupervisedQA}} to re-implement their approach based on BERT-Large (Whole Word Masking) model.

Table \ref{tab:mainresult} shows the main results on SQuAD 1.1 and NewsQA. Training QA model on our~\refqa{} outperforms the previous methods by a large margin.
Combining with iterative data refinement, our approach achieves new state-of-the-art results in the unsupervised setting.
Our QA model attains 71.4 F1 on the SQuAD 1.1 test set and 45.1 F1 on the NewsQA test set without using their annotated data, outperforming all of the previous unsupervised methods.
In particular, the results are competitive with early supervised models.

\subsection{Analysis}

We conduct ablation studies on the SQuAD 1.1 dev set, in order to better understand the contributions of different components in our method.

\subsubsection{Effects of \refqa{}}

%\citet{lewis2019unsupervisedqa} generate questions from the sentences or sub-clauses of the paragraph in Wikipedia. Different from their approach, we generate questions from statements in Wikipedia and use it cited documents as our context.
We conduct experiments on \refqa{} and another synthetic dataset (named as \textsc{Wiki}). The \textsc{Wiki} dataset is constructed using the same method as in~\citet{lewis2019unsupervisedqa}, which uses Wikipedia pages as context paragraphs for QA examples.
% \zhongli{The main difference between the two datasets is that the contexts of \refqa{} are from Wikipedia references, but the \textsc{Wiki} contexts are from Wikipedia paragraphs.}
In addition to the dependency reconstruction method (Section~\ref{sec:dep:rec}), we compare three cloze translation methods proposed in~\citet{lewis2019unsupervisedqa}.

\begin{table}[t]
\small
\centering
\begin{tabular}{p{0.8cm}p{1.1cm}<{\centering}p{1.1cm}<{\centering}p{1.1cm}<{\centering}p{1.2cm}<{\centering}}
%\begin{tabular}{p{1.2cm}p{1.35cm}p{1.35cm}p{1.35cm}p{1.35cm}}
\toprule
& Identity & Noise & UNMT & DRC \\
\midrule
%Wiki & 30.5 & 45.6 & 49.1 &  35.7 \\
%\refqa{} & \textbf{51.6} &  \textbf{53.5} 
%        &  \textbf{52.0} &  \textbf{58.8} \\
\textsc{Wiki} & 20.8~/~30.5 & 36.6~/~45.6 & 40.5~/~49.1 & 26.3~/~35.7 \\
\refqa{} & \textbf{42.5}~/~\textbf{51.6} &  \textbf{45.1}~/~\textbf{53.5} 
& \textbf{43.4}~/~\textbf{52.0} & \textbf{49.2}~/~\textbf{58.8} \\
\bottomrule
\end{tabular}
\caption{Results (EM / F1) of \refqa{} and \textsc{Wiki} datasets with different cloze translation methods on the SDuAD 1.1 dev set. ``DRC" is short for dependency reconstruction.}
\label{tab:ref&top}
\end{table}

\begin{table}[t]
\small
\centering
\begin{tabular}{p{3.2cm}|p{3.2cm}}
\toprule
it finished first in the \textcolor{blue}{Arbitron} ratings in April 1990 & he was sold to Colin Murphy's Lincoln City for a fee of \textcolor{blue}{15,000} \\ 
\midrule
\textbf{UNMT:} Who finished it first in the ratings in April 1990 ? & \textbf{UNMT:} How much do we need Colin Murphy 's Lincoln City for a fee ? \\ 
\textbf{DRC:} Who ratings in it finished first in April 1990 & \textbf{DRC:} How much of a fee for he was sold to Colin Murphy 's Lincoln City \\
\bottomrule
\end{tabular}
    %\begin{tabular}{cp{5.5cm}}
    %\toprule
    %    Statement & Bierut is the co-founder of the blog \textcolor{blue}{DesignObserver} \\
    %    UNMT &  What is the Bierut co-founder of the blog ? \\
    %    DRC  &  What of the co-founder Bierut is \\ \midrule
    %    Statement & he was sold to Colin Murphy's Lincoln City for a fee of \textcolor{blue}{15,000} \\
    %    UNMT & How much do we need Colin Murphy 's Lincoln City for a fee ?\\
    %    DRC & How much of a fee for he was sold to Colin Murphy 's Lincoln City \\
    %\bottomrule
    %\end{tabular}
    \caption{Examples of generated questions using UNMT and our method. ``DRC" is short for our dependency reconstruction. The blue words indicate extracted answers.}
    \label{tab:example_q}
\end{table}

\begin{table}[t]
\small
\centering
\begin{tabular}{p{3.3cm}p{0.4cm}<{\centering}p{0.9cm}<{\centering}p{1.3cm}<{\centering}}
\toprule
 & Iter.  &  Size  & EM / F1   \\ \midrule
Initial QA Model  & &  300k       & 57.1~/~66.8 \\ \midrule
Training on \\
\quad Filtered Data & \xmark & 464k & 57.4~/~67.1 \\
\quad Refined Data & \xmark & 100k  & 61.0~/~70.7 \\ 
\quad Refined + Filtered Data & \xmark & 200k & 61.8~/~71.0 \\
\quad Refined Data & \cmark & $\text{6}\times\text{15k}$ & 60.1~/~70.0 \\
\quad  Refined + Filtered Data & \cmark  & $\text{6}\times\text{30k}$ & \textbf{62.5}~/~\textbf{72.6} \\ 
% \hline
% Retrain & no & 358k   & 58.4 & 69.2 \\
\bottomrule
\end{tabular}
\caption{Results of using filtered data, refined data, and the combination for data refinement on the SDuAD 1.1 dev set. ``Iter." is short for iterative training.}
\label{tab:merge}
\end{table}

\paragraph{Identity Mapping} generates questions by replacing the mask token in cloze questions with a relevant wh* question word.
% shown in Table~\ref{tab:qmap}.

\paragraph{Noise Cloze} first applies a noise model, such as permutation, and word drop, as in~\citet{lample2017unsupervised}, and then applies the ``Identity Mapping" translation.

\paragraph{UNMT} converts cloze questions into natural questions following unsupervised neural machine translation.
%trained on non-aligned corpora of natural questions and cloze questions. 
Here we directly use the published model of~\citet{lewis2019unsupervisedqa} for evaluation.

%\wang{Besides, we compare our method of translating clozes to questions with other unsupervised question generation methods:} 

%\zhongli{To compare with our \refqa{}, we obtain the top 10,000 articles of English Wikipedia, create a synthetic dataset (named as ``Wiki") by following the methodology of~\citet{lewis2019unsupervisedqa}.}
For a fair comparison, we sample 300k training data for each dataset, and fine-tune BERT-Base for 2 epochs.
% conduct experiments by sampling 300k training data for two dataset, fine-tuning BERT-base with 2 epochs and evaluating on SQuAD 1.1 dev set.
%Different from~\citet{lewis2019unsupervisedqa}, which generates questions from the same paragraphs, we generate questions from statements in Wikipedia and use its cited documents as our paragraphs. 
As shown in Table~\ref{tab:ref&top}, training on our~\refqa{} achieves a consistent gain over all cloze translation methods.
%, and our translation method significantly improves the performance on \refqa{}. 
Moreover, our dependency reconstruction method is also favorable compared with the ``Identity Mapping" method.
%is based on the method ``identity". The results on ``Wiki" suggest that our method is still useful owing to 5.2 absolute improvement comparing to ``identity".
%Comparing to other translation methods, our method achieves the best performance, outperforming other methods with 5.3 absolute improvement on \refqa.
The improvement of DRC on \textsc{Wiki} is smaller than on \refqa{}. We argue that it is because \textsc{Wiki} contains too many lexical overlaps, while DRC mainly focuses on providing structural diversity.

We present the generated questions of our method (DRC) and UNMT in Table~\ref{tab:example_q}. Most natural questions follow a similar structure: question word (what/who/how), question focus (name/money/time), question verb (is/play/take) and topic~\citep{qa:as:ie}. Compared with UNMT, our method adjusts answer-related words in the dependency tree according to the linguistic characteristics of natural questions.

%\subsubsection{Effect of Translation Method}
%We compare our syntax-guided translation to other unsupervised question generation methods. 

%Table~\ref{tab:ref&top} shows the results of different question generation methods. Our syntax-guided translation with data augmentation achieves the best performance (58.8 F1 score), outperforming other methods with 5.3 absolute improvement on \refqa. Moreover, our syntax-guided translation can be treated as a combination of ``identity" and the dependency tree reconstruction. The results on ``Wiki" suggest that our method is still useful owing to 5.2 absolute improvement comparing to ``identity".

% for Data Refinement
\subsubsection{Effects of Data Combination}
%New data can be generated through our data refinement and there are many approaches of using refined data. 

We validate the effectiveness of combining refined and filtered data for our data refinement.
We use only refined or filtered data to train our QA model, comparing with the combining approach.

The results are shown in Table~\ref{tab:merge}. We observe that both data can help the QA model to achieve better performance. Moreover, the combination of refined and filtered data is more useful than only using one of them.
Using iterative training, our combination approach further improves the model performance to 72.6 F1 (1.6 absolute improvement).
%Training on the refined and filtered data, our iterative training approach achieves the best performance with 72.6 F1 (1.6 absolute improvement). 
Besides, using our refined data contributes further improvement compared with filtered data.
%suggesting that there may be more appropriate and diverse answers found by our data refinement.
%Training on \refqa, we have achieved 66.8 F1 on SQuAD 1.1 dev set, and then we conduct experiments on the following settings for data refinement:

%\textbf{Retrain} We discard the current model, combine new data with the previous training data, and retrain our QA model.

%\textbf{Original Data} We discard the refined data, select the original data which answers satisfy the probability threshold $\tau$ as new training data to fine-tune the current model. 

%\textbf{Refined Data} We fine-tune the current model by using refined data directly.

%\textbf{Refined + Original Data} We combine refined data with original data in equal proportion, and then fine-tune the current model.

%As shown in Table~\ref{tab:merge}, the combination of refined and original data achieves better performance than using only refined or original data. 
%and the multi-round setting helps to achieve the best performance.
%Moreover, using our refined data achieves at least 3.2 improvement, that shows robustness of our data refinement.

\begin{table}[t]
    \small
    \centering
    \begin{tabular}{cccccccc}
        \toprule
        $\tau$ & 0.0 & 0.1 & 0.15 & 0.2 & 0.3 & 0.5 & 0.7  \\ \midrule
        EM & 54.3 & 61.2 & 61.8 & 61.1 & 59.7 & 59.2 & 58.5 \\
        F1 & 69.6 & 70.4 & 71.0 & 70.9 & 69.4 & 68.7 & 67.7 \\
        \bottomrule
    \end{tabular}
    \caption{Results of using different confidence thresholds during the construction of the refined data and filtered data.}
    \label{tab:threshold}
\end{table}

\begin{figure}[t]
\centering
\includegraphics[width=\linewidth]{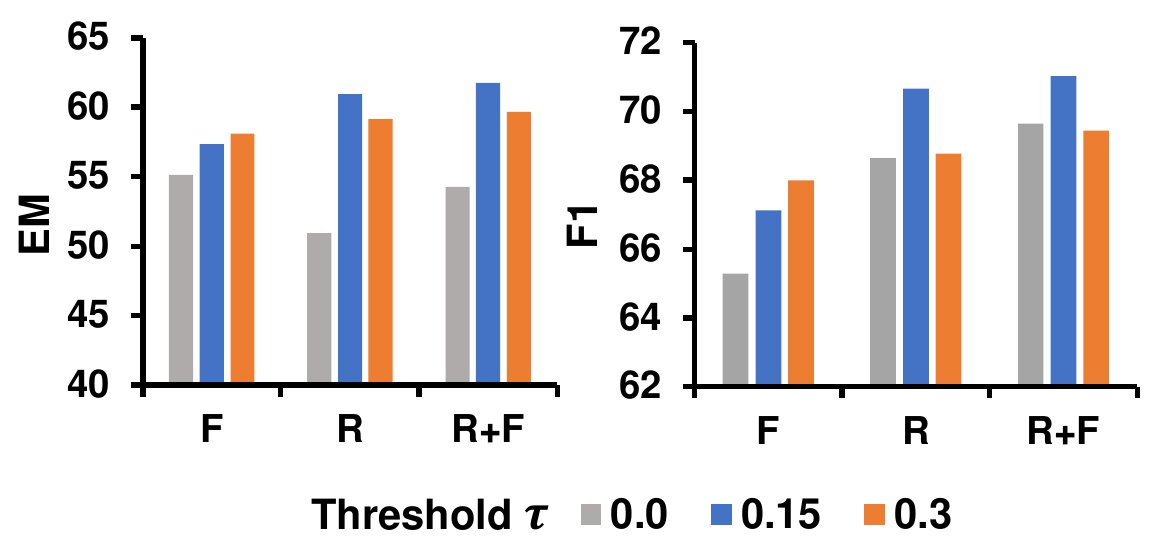}
\caption{Comparison on filtered data and refined data with different confidence thresholds. ``F" is short for using filtered data, ``R" is short for using refined data. ``R+F" is short for the combination of refined and filtered data.}
\label{fig:threshold}
\end{figure}

\subsubsection{Effects of Confidence Threshold}
We experiment with several thresholds (0.0, 0.1, 0.15, 0.2, 0.3, 0.5 and 0.7) to filter the predicted answers. Their QA results on SQuAD 1.1 dev set are presented in Table~\ref{tab:threshold}. Using threshold of 0.15 achieves better performance. 

% analyze the confidence threshold $\tau$ by filtering the predicted answers with several thresholds (0.0, 0.1, 0.15, 0.2, 0.3, 0.5 and 0.7). The final threshold 0.15 is chosen based on the QA performance shown in Table~\ref{tab:threshold}.}

We also analyze the effects of threshold on refined data and filtered data.
%We conduct experiments in the above non-iterative settings. 
As shown in Figure~\ref{fig:threshold}, for the filtered data, using a higher confidence threshold achieves better performance, suggesting that using the QA model for filtering makes our examples more credible.
For the refined data and the combination, we observe that the threshold 0.15 achieves a better performance than the threshold 0.3, but the EM is greatly reduced when the threshold is set to 0.0. Besides, there are 26,257 answers that do not appear in named entities using the threshold 0.15, compared to 15,004 for the threshold 0.3. Thus, an appropriate threshold can help us improve the answer diversity and get rid of some noisy examples.

%\zhongli{This shows an appropriate threshold can help us improve the answer diversity and get rid of some noisy examples.}

\begin{comment}
\subsubsection{Effects of Iterative Training}
Table~\ref{tab:merge} shows that combining the refined data with filtered data, iterative training achieves the best performance with 72.6 F1 on the SQuAD 1.1 dev set. As shown in Figure~\ref{fig:multiround}, we also investigate the performance of iterative training with respect to the number of iterations.

%For iterative setting, the remaining data are split into 6 parts to do refinement on each iteration. 
%For each round, we use the previous best model to refine one part question-answer pairs and . 
%Figure~\ref{fig:multiround} shows the results on each round for different above settings.

\begin{figure}[t]
    \centering
    \includegraphics[width=7.5cm]{images/multiround.pdf}
    \caption{Performance on each iteration for the settings of using only refined data and combining with filtered data on the SQuAD 1.1 dev set.}
    \label{fig:multiround}
\end{figure}
\end{comment}

\subsubsection{Effects of Refinement Types}
\label{sec:eda}

For brevity, we denote the original answer and predicted answer by ``OA" and ``PA", respectively.
In order to analyze the contribution of our refined data, we categorize the data refinements into the following three types:
% Besides, the following predicted answers are different from the original answers: 

\noindent
\textbf{OA$\bm{\supset}$PA} The original answer contains the predicted answer.

%\textbf{$\text{OA }\mathbf{\subset}\text{ PA}$}
\noindent
\textbf{OA$\bm{\subset}$PA} The predicted answer contains the original answer.

\noindent
\textbf{Others} The remaining data except for the above two types of refinement.

\begin{table}[t]
% \small
\centering
\begin{tabular}{lcccc}
\toprule
  & Refined & Size  & EM / F1   \\ \midrule
\refqa{} & - & 300k & 57.1 / 66.8 \\ \midrule
% All Data& 200k & \textbf{61.8} & \textbf{71.0} \\ \midrule 
OA$\bm{\supset}$PA & \xmark & 90k & 59.4 / 69.0 \\
OA$\bm{\supset}$PA & \cmark & 90k & 50.9 / 64.6 \\
OA$\bm{\subset}$PA & \xmark & 35k & 47.5 / 61.2 \\
OA$\bm{\subset}$PA & \cmark & 35k & 60.3 / 69.9 \\
Others & \xmark& 75k & 52.2 / 62.3 \\
Others & \cmark& 75k & 58.8 / 69.7 \\
\bottomrule
\end{tabular}
\caption{Comparison between different types of data refinement on the SQuAD 1.1 dev set.}
\label{tab:case}
\end{table}

\begin{table*}[t!]
% \small
\centering
\begin{tabular}{lp{13cm}}
\toprule
\textbf{OA$\bm{\supset}$PA} 
& \textbf{S:} In 1938, E. Allen Petersen escaped the advancing Japanese armies by sailing a junk, ``Hummel Hummel", from Shanghai to California with his wife Tani and two White Russians (Tsar loyalists).  \\
& \textbf{Q:} Who escaped the advancing Japanese armies by sailing a junk   \\
& \textbf{OA:} E. Allen Petersen   \\
& \textbf{PA:} Petersen \\
& \textbf{RQ:} Who escaped the advancing Japanese armies by sailing a junk  \\ \midrule
\textbf{OA$\bm{\subset}$PA}
& \textbf{S:} Hyundai announced they would be revealing their future rally plans at the 2011 Chicago Auto Show on February 9 .  \\
& \textbf{Q:} What at they would be revealing their future rally plans on February 9  \\
& \textbf{OA:} Chicago Auto Show  \\
& \textbf{PA:} the 2011 Chicago Auto Show \\
& \textbf{RQ:} What at their future rally plans they would be revealing on February 9 \\ \midrule
\textbf{Others}
& \textbf{S:} In January 2017, she released the track ``That's What's Up" that re-imagines the spoken word segment on the Kanye West song ``Low Lights". \\
& \textbf{Q:} What the Kanye West song on the spoken word segment re-imagines  \\
& \textbf{OA:} Low Lights  \\
& \textbf{PA:} That's What's Up  \\
& \textbf{RQ:} What the track she released that re-imagines the spoken word segment on the Kanye West song ``Low Lights" . \\
\bottomrule
\end{tabular}
\caption{The generated and refined question-answer pairs. ``S" and ``Q" are short for statement and question. ``OA", ``PA" and ``RQ" are short for the original answer, predicted answer and the refined question.}
\label{tab:examples}
\end{table*}

For each type, we keep the original data or use refined data to train our QA model.
%The refined triples are generated by using the predicted answers as new answer candidates. 
We conduct experiments on the non-iterative setting with the data combination.

As shown in Table~\ref{tab:case}, our refined data improves the QA model in most types of refinement except ``OA$\bm{\supset}$PA".
The results indicate that the QA model favors longer phrases as answer spans.
% As shown in , experimental results show all cases outperform the previous model consistently, suggesting that each case may help improve the quality of question-answer pairs.
%suggesting that each case may help improve the diversity of questions and answers. 
%In particular, both ``OA$\bm{\supset}$PA" and ``Others" can achieve a similar result of using all data.
Moreover, for the ``OA$\bm{\subset}$PA" and ``Others" types, there are 47.8\% answers that are not extracted by the NER toolkit.
The iterative refinement extends the category of answer candidates, which in turn produces novel question-answer pairs.

%\subsubsection{Quality of Generated Data}
We show a few examples of our generated data in Table~\ref{tab:examples}.
We list one example for each type.
For the ``OA$\bm{\supset}$PA" refinement, the predicted answer is a sub-span of the extracted named entity, but the complete named entity is more appropriate as an answer.
For the ``OA$\bm{\subset}$PA" refinement, the QA model can help us extend the original answer to be a longer span, which is more complete and appropriate.
Besides, for the ``Others" refinement, its prediction can be a new answer, and not appear in named entities extracted by the NER toolkit. 
%which helps us to extend the category of answer candidates.

%We observe that the QA model predict wrong answers for our first translated questions such as the ``OA$\bm{\supset}$PA" and ``Others" cases. For the ``OA$\bm{\supset}$PA" case, \wang{we keep the original answer and question because of our beliefs that the extracted named entity is indivisible.} But for the ``Others" case, its prediction can be a new answer despite it's wrong for the input question. 
%Besides, the extracted answers are not always suitable by using the NER tool for the ``OA$\bm{\subset}$PA" case. Thanks for BERT linguistic pre-training, our QA model can help us to find more appropriate answers.

%\zhongli{
%\subsubsection{Quality of Generated Question}
%Most questions follow a similar structure: question word (what/who/how), question focus (name/money/time), question verb (is/play/take) and topic~\citep{qa:as:ie}.
%As shown in Table~\ref{tab:example_q}, our method adjusts answer-related words in the dependency tree according to the linguistic characteristics of natural questions. However, the UNMT method makes drifts on the question focus and topic, though its questions are more grammatic and fluent.
%}

%Moreover, from the questions in Table~\ref{tab:examples}, we see that our translation method makes answer-related words in front, specially for the questions started with ``\textit{what}" and ``\textit{how many/much}". But the fluency of generated questions is not good obviously.

\subsection{Few-Shot Learning}

Following the evaluation of~\cite{yang-semiqa,dhingra2019semi}, we conduct experiments in a few-shot learning setting. 
We use the best configuration of our approach to train the unsupervised QA model based on BERT-Large (Whole Word Masking).
Then we fine-tune the model with limited SQuAD training examples.

\begin{figure}[t]
\centering
\includegraphics[width=\linewidth]{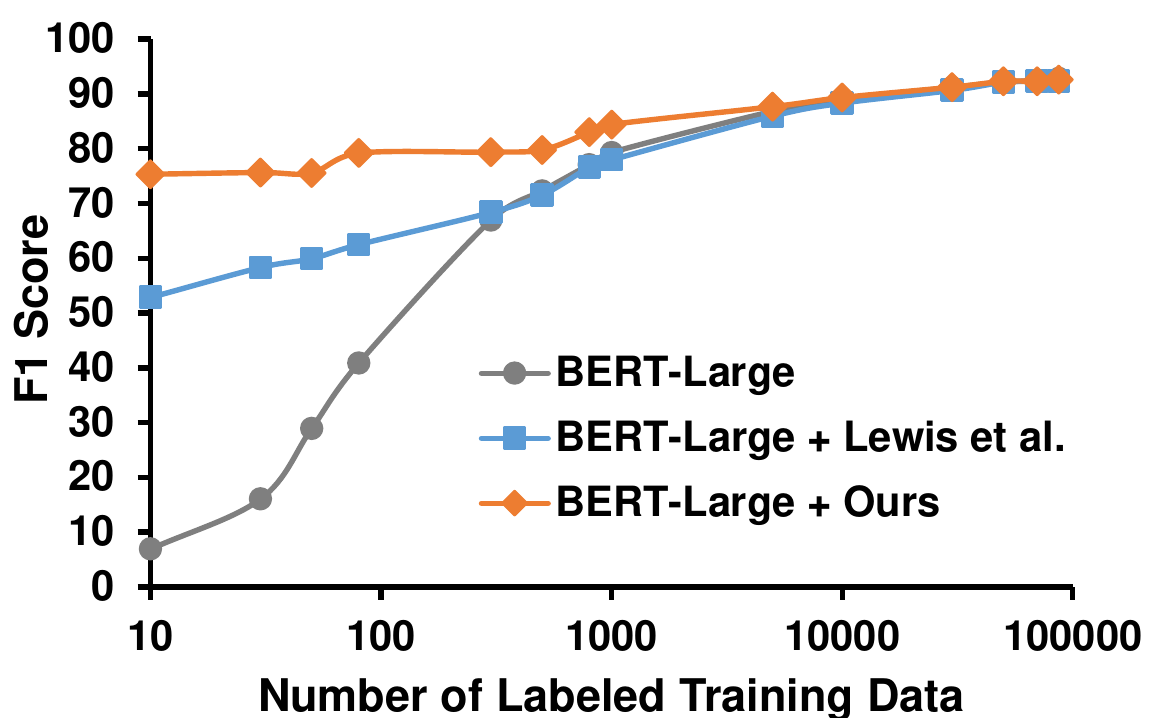}
\caption{F1 score on the SQuAD 1.1 dev set with various training dataset sizes.}
\label{fig:fewshot}
\end{figure}

As shown in Figure~\ref{fig:fewshot}, our method obtains the best performance in the restricted setting, compared with the previous state of the art~\cite{lewis2019unsupervisedqa} and directly fine-tuning BERT.
Moreover, our approach achieves $79.4$ F1 (16.4 absolute gains than other models) with only $100$ labeled examples.
The results illustrate that our method can greatly reduce the demand of in-domain annotated data.
In addition, we observe that the results of different methods become comparable when the labeled data size is greater than 10,000.

\section{Conclusion}

In this paper, we present two approaches to improve the quality of synthetic QA data for unsupervised question answering. We first use the Wikipedia paragraphs and its references to construct a synthetic QA data \refqa{} and then use the QA model to iteratively refine data over \refqa{}. Our method outperforms the previous unsupervised state-of-the-art models on SQuAD 1.1, and NewsQA, and achieves the best performance in the few-shot learning setting. 
%In this paper, we first take advantage of the natural difference in expression from the statements and references in Wikipedia, and transform these to a synthetic augmented dataset automatically. The construction of augmented dataset makes our generated questions divergent from context. Then, we observe that the answers extracted by pretrained components are not suitable for context in some cases, thus we develop a data refinement approach to use the QA model to help us extract more appropriate answers. The results on SQuAD 1.1 benchmark show that our approaches improve the unsupervised QA performance consistently, and the final result achieves the state-of-the-art. As for future work, we would like to design models to infer answer types and generate more fluent questions for unsupervised QA. 

\section*{Acknowledgements}
The work was partially supported by National Natural Science Foundation of China (NSFC) [Grant No. 61421003].

\bibliography{uqa}
\bibliographystyle{acl_natbib}

\end{document}